\documentclass[10pt,twocolumn,letterpaper]{article}

\usepackage[pagenumbers]{cvpr} 

\usepackage{graphicx}
\usepackage{amsmath}
\usepackage{amssymb}
\usepackage{booktabs}

\usepackage{multirow}
\usepackage[american]{babel} 
\graphicspath{{./Figure/}}
\usepackage{pifont}
\newcommand{\cmark}{\ding{51}}

\newcommand{\squishlist}{
 \begin{list}{$\bullet$}
  { \setlength{\itemsep}{0pt}
     \setlength{\parsep}{1pt}
     \setlength{\topsep}{1pt}
     \setlength{\partopsep}{0pt}
     \setlength{\leftmargin}{1em}
     \setlength{\labelwidth}{1em}
     \setlength{\labelsep}{0.5em} } }

\newcommand{\squishend}{
  \end{list}  }

\usepackage[pagebackref,breaklinks,colorlinks]{hyperref}

\def\cvprPaperID{8072} 
\def\confName{CVPR}
\def\confYear{2022}

\begin{document}

\title{Compressing Models with Few Samples: Mimicking then Replacing}

\author{Huanyu Wang$^{1}$\thanks{Corresponding author.}, Junjie Liu$^{2}$, Xin Ma$^{2}$, Yang Yong$^{2}$, Zhenhua Chai$^{2}$, Jianxin Wu$^{1}$\\
\normalsize$^1$ State Key Laboratory for Novel Software Technology, Nanjing University\\
\normalsize$^2$ Meituan\\
{\tt\small \{cjnjuwhy,wujx2001\}@gmail.com, \{liujunjie10,maxin10,chaizhenhua\}@meituan.com}
}
\maketitle

\begin{abstract}
Few-sample compression aims to compress a big redundant model into a small compact one with only few samples. If we fine-tune models with these limited few samples directly, models will be vulnerable to overfit and learn almost nothing. Hence, previous methods optimize the compressed model layer-by-layer and try to make every layer have the same outputs as the corresponding layer in the teacher model, which is cumbersome. In this paper, we propose a new framework named Mimicking then Replacing (MiR) for few-sample compression, which firstly urges the pruned model to output the same features as the teacher's in the penultimate layer, and then replaces teacher's layers before penultimate with a well-tuned compact one. Unlike previous layer-wise reconstruction methods, our MiR optimizes the entire network holistically, which is not only simple and effective, but also unsupervised and general. MiR outperforms previous methods with large margins. Codes will be available soon.
\end{abstract}

\section{Introduction}\label{sec:intro}

Convolutional neural networks (CNNs) with millions of parameters can only be utilized by high-performance devices, even when we only care about the inference stage. In order to put deep models into small devices and decrease the latency and memory consumption, network compression~\cite{deng_compression_survey_2020} is widely used in model deployment. To compress a model, network pruning methods~\cite{he_channel_2017, luo_thinet_2017, zhu2017prune, li_eagleeye_2020, luo_neural_2020,liu_fisherpruning_2021} try to prune less useful weights or channels, while quantization methods~\cite{gholami2021survey} aim at quantizing the weights and activations with fewer bits, and knowledge distillation methods~\cite{hinton_kd_2015, Romero_fitnets_2015} try to distill the dark knowledge from a potentially redundant big model into a more compact small one. 

These compression methods have been very successful in reducing computations and accelerating inference speed. But, they all assume full access to the training data, and unfortunately this assumption does not hold in many cases, especially in non-academic scenarios. When handling sensitive data (\eg, medical or commercial data), data security issues are of special importance. As a response to this issue, the few-sample or few-shot compression problem aims to compress models with limited samples, which is a practical way to protect data privacy by using only non-sensitive data.

To tackle this few-sample compression problem, recent methods try to obtain a compact model \emph{in a layer-wise manner}. Li \etal proposed FSKD~\cite{Li_2020_CVPR}, which adds a $1 \times 1$ conv. after each layer and optimizes the weights by minimizing the reconstruction error for each layer. Bai \etal introduced a cross distillation operation to better alleviate the error accumulation in each layer~\cite{Bai_Wu_King_Lyu_2020}. Furthermore, Shen \etal tried to distill a compact model by grafting layers from the teacher model to the student model progressively~\cite{Shen_Wang_Yin_Song_Luo_Song_2021}. All these methods tried to reconstruct the representation ability layer-wisely, which is not only cumbersome but may also cause \emph{error accumulation}. Moreover, this layer-wise framework needs a one-to-one relationship between the pruned and the original model, therefore imposing \emph{heavy restrictions} to the pruned model's structure.

\begin{figure*}
	\centering
	\includegraphics[width=0.95\textwidth]{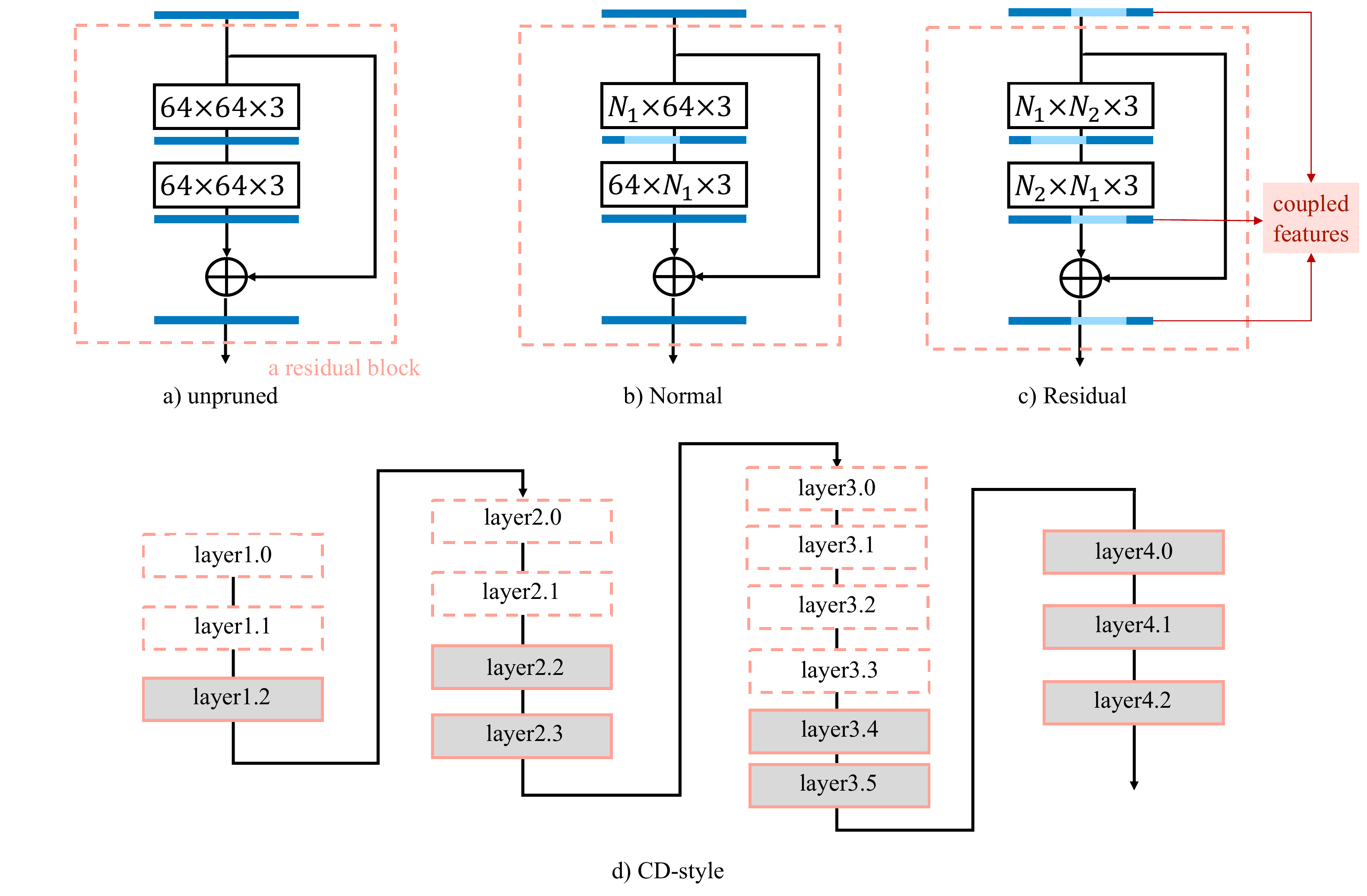}
	\caption{Illustration of different pruning schemes. We use a rectangle with notation $N_1\times N_2\times K$ to represent a conv. with $N_1$ output channels, $N_2$ input channels, and kernel size $K$. Blue rectangles represent features (activation maps). In this figure: a) A residual block in ResNet-34 contains two conv., and we omit batch normalization and non-linear layers; b) In the `Normal' pruning scheme, only channels within residual blocks are pruned, in which the light blue color indicates channels that are pruned; c) the `Residual' pruning scheme, which not only prunes channels within blocks, but also prunes coupled channels across different residual blocks; d) In the `CD' pruning scheme, only dashed and transparent residual blocks are pruned, and these blocks are pruned with the `Normal' scheme. Best viewed in color.}
	\label{fig1}
\end{figure*}

Instead, we advocate pruning and optimizing the entire network \emph{holistically} instead of following this cumbersome layer-wise reconstruction framework, and recover the representation abilities of the pruned model \emph{globally} instead of training the layers locally. In this regard, we propose a new framework, Mimicking then Replacing (MiR), which first urges the pruned model to output the same features as the teacher's (\ie, \emph{mimicking} features), and following LSHKD~\cite{wang_mimicking_2021} we can mimic the features in the penultimate layer. Then, while keeping the (classification, detection, \etc) head intact, we \emph{replace} all the other layers with the trained compact model after mimicking. The features in the penultimate layer in LSHKD~\cite{wang_mimicking_2021} are obtained after a pooling layer. We reveal that mimicking the features \emph{before} the pooling layer can boost the accuracy without extra computation.  

MiR is simple to use (simple algorithm and zero extra hyperparameters), general (suitable for many scenarios), unsupervised (not even use labels for the few-sample training set) and highly accurate (outperforming current state-of-the-art methods by a large margin). With MiR, we can train an accurate compressed model within dozens of minutes and only hundreds of samples. To sum up, our contributions are:
\squishlist
  \item We propose a simple but effective framework, Mimicking then Replacing (MiR), for few-sample compression. MiR contains no extra hyperparameters to tune but outperforms state-of-the-art methods with a large margin.
  \item MiR is general to use. It can be used for different pruning schemes and is effective in different network architectures. Moreover, it has no restriction on the model's structure, and can avoid error accumulation because we are the first to optimize the weights holistically in few-shot compression.
  \item LSHKD~\cite{wang_mimicking_2021} mimics the features in the penultimate layer, but we find that mimicking features \emph{before} pooling helps a lot, at least in few-shot compression. It brings 0.5 to 2.0 percentage points without extra computation compared with mimicking features after the final pooling layer (\ie, the penultimate layer).
\squishend

\section{Related Work}

\noindent\textbf{Pruning.} 
Network pruning is an effective and general way to reduce model size and computations~\cite{han2015learning, choudhary2020comprehensive, he2018soft}. Exiting pruning works can be divided into two categories: unstructured pruning and structured pruning. Unstructured pruning aims to prune connections, leading to unstructured sparsity of models. Han \etal proposed a three-step method to prune redundant connections \cite{han2015learning}. Dynamic network surgery designed by Guo \etal can integrate connection splicing into the pruning process, which can significantly reduce network complexity~\cite{guo2016dynamic}. Tung and Mori proposed CLIP-Q, which combined the advantages of weight pruning and weight quantization in a single framework~\cite{tung2018clip}. 

Structured pruning tries to prune less useful channels and is friendly to all platforms. Li \etal first proposed to accelerate CNNs by removing filters that have smaller $\ell_1$-norm~\cite{li_pruning_2017}. Luo \etal pruned channels based on the statistics computed from the next layer \cite{luo_thinet_2017}. And there are other methods~\cite{lin2020hrank, dubey2018coreset, he2018soft,li_eagleeye_2020,liu_fisherpruning_2021} that tried different ways to define the importance of channels. Instead of finding the importance manually, \cite{liu_metapruning_2019, chen_network_2020, guo_dmcp_2020} tried to automatically find better pruned models. Luo and Wu~\cite{luo_neural_2020} pruned the residual connections to get wallet-shaped models, which have both higher inference speed and higher accuracy.

\noindent\textbf{Knowledge distillation (KD).} 
KD distills knowledge from a redundant well-trained model into a smaller model, and most KD methods focus on finding better knowledge or a better way to distill knowledge. Hinton \etal first adopted KD and tried to distill from the softmax outputs~\cite{hinton_kd_2015}. FitNet~\cite{Romero_fitnets_2015} used not only the outputs but also the intermediate representations as hints to guide the student. Recently, Wang \etal argued that it is better to \emph{only} use the teacher's features in the penultimate layer and added an LSH loss to make the student focus more on the feature directions and less on magnitudes. We follow LSHKD~\cite{wang_mimicking_2021} to only use the final features in distillation.

\noindent\textbf{Compression with limited data.} 
Previous few-shot compression methods mostly compress a network layer by layer. Li \etal proposed FSKD \cite{Li_2020_CVPR}, which contains three steps to train a pruned layer. The first step is to add a $1 \times 1$ conv. layer after each layer, the second step calculates the weights in the $1 \times 1$ conv. by solving a least-square problem, and the last step merges the $1 \times 1$ conv. into the original conv. layer. Bai \etal proposed a cross distillation operation in CD \cite{Bai_Wu_King_Lyu_2020}, which includes correction and imitation, then optimized the compressed model layer-by-layer, cross distilled by the guidance of a pre-trained model. Shen \etal proposed a grafting method to align each layer by layer-wisely grafting the student's layer into the teacher's, and by minimizing the estimation error of the output logits~\cite{Shen_Wang_Yin_Song_Luo_Song_2021}.

There are also methods on zero-shot compression. Chen \etal proposed a data-free method DAFL, which treated pre-trained teacher networks as the discriminators and trained a generator for deviating training samples~\cite{Chen_2019_ICCV}. Haroush \etal generated synthetic samples for calibrating and fine-tuning quantized models without any real data~\cite{Haroush_2020_CVPR}.  

In this paper, we mainly compare our MiR with two pruning-based methods FSKD~\cite{Li_2020_CVPR} and CD~\cite{Bai_Wu_King_Lyu_2020}.

\begin{table*}
	\footnotesize
	\setlength\tabcolsep{1.325pt}
	\caption{Mean and standard deviation of top-1/top-5 accuracy (\%) on ILSVRC-2012. We used `Prune-C Normal' to prune ResNet-34 and compared different methods with different training sizes. We used 50, 100, 500 random samples, and $N$-way-$K$-shot ($N/K$ in the top row) settings. All the results were reported with five trails.}
	\centering
	\begin{tabular}{cccc|ccc}
		\toprule
		Method	&	50			&	100			&	500			&	1000/1			&	1000/2			&	1000/3			\\
		\midrule																									
		BP	&	$39.0_{\pm 1.41}	/	{68.9_{\pm 1.17}}$	&	$41.0_{\pm 0.33}	/	{70.5_{\pm 0.66}}$	&	$51.8_{\pm 0.30}	/	{78.1_{\pm 0.38}}$	&	$57.8_{\pm 0.30}	/	{81.5_{\pm 0.18}}$	&	$60.0_{\pm 0.23}	/	{83.0_{\pm 0.11}}$	&	$61.0_{\pm 0.19}	/	{83.7_{\pm 0.15}}$	\\
		KD	&	$44.5_{\pm 1.20}	/	{72.3_{\pm 0.87}}$	&	$46.4_{\pm 0.34}	/	{74.0_{\pm 0.58}}$	&	$54.7_{\pm 0.26}	/	{79.7_{\pm 0.19}}$	&	$57.9_{\pm 0.21}	/	{81.6_{\pm 0.12}}$	&	$59.0_{\pm 0.14}	/	{82.4_{\pm 0.15}}$	&	$59.3_{\pm 0.07}	/	{82.6_{\pm 0.08}}$	\\
		FSKD	&	$45.3_{\pm 0.77}	/	{71.5_{\pm 0.62}}$	&	$51.2_{\pm 0.30}	/	{76.8_{\pm 0.23}}$	&	$57.6_{\pm 0.21}	/	{81.6_{\pm 0.15}}$	&	$59.4_{\pm 0.13}	/	{82.7_{\pm 0.06}}$	&	$60.1_{\pm 0.13}	/	{83.2_{\pm 0.08}}$	&	$60.3_{\pm 0.12}	/	{83.4_{\pm 0.05}}$	\\
		CD	&	$56.2_{\pm0.37}	/	{80.8_{\pm 0.31}}$	&	$59.1_{\pm0.22}	/	{82.8_{\pm 0.11}}$	&	$63.7_{\pm0.18}	/	{86.0_{\pm 0.05}}$	&	$64.4_{\pm0.03}	/	{86.3_{\pm 0.07}}$	&	$64.9_{\pm0.13}	/	{86.6_{\pm 0.08}}$	&	$65.2_{\pm0.09}	/	{86.7_{\pm 0.07}}$	\\
		\midrule
		$\text{MiR}_{after}$	&	$61.0_{\pm0.21}	/	{84.3_{\pm0.16}}$	&	$62.5_{\pm0.17}	/	{85.4_{\pm0.13}}$	&	$65.4_{\pm0.03}	/	{87.2_{\pm0.13}}$	&	$66.6_{\pm0.06}	/	{87.8_{\pm0.06}}$	&	$67.2_{\pm0.10}	/	{88.1_{\pm0.04}}$	&	$67.5_{\pm0.07}	/	{88.3_{\pm0.06}}$	\\
		$\text{MiR}_{before}$	&	$64.1_{\pm0.10}	/	{86.3_{\pm0.11}}$	&	$65.1_{\pm0.19}	/	{87.0_{\pm0.11}}$	&	$67.0_{\pm0.09}	/	{88.1_{\pm0.07}}$	&	$67.8_{\pm0.06}	/	{88.5_{\pm0.02}}$	&	$68.2_{\pm0.10}	/	{88.8_{\pm0.04}}$	&	$68.4_{\pm0.09}	/	{88.9_{\pm0.02}}$	\\
		\bottomrule
	\end{tabular}
	\label{tab:normal_pruning}
\end{table*}


\section{The Propose MiR Method}

In this section, we first look back at the layer-wise reconstruction framework for few-sample compression. Then, we point out the shortcomings of this framework, and propose a new framework called Mimicking then Replacing (MiR) to deal with this few-sample compression problem.

First of all, we define the notation which are used in this paper. We aim to get a pruned model $\mathcal{M}_{P}$ from the original pre-trained model $\mathcal{M}_{O}$, and with the few-sample data $\mathcal{D}_{few} \subset \mathcal{D}_{train}$, where $\mathcal{D}_{train}$ is the original training dataset. In some cases, even labels for $\mathcal{D}_{train}$ may not be available, \ie, we prefer an unsupervised compression.

The original model $\mathcal{M}_{O}$ is well-trained using the full dataset $\mathcal{D}_{train}$. The weights of the $l$-th layer in $\mathcal{M}_{P}$ and $\mathcal{M}_{O}$ are denoted as $W^{l}_{P}$ and $W^{l}_{O}$, respectively. Similarly, the features (activation maps) are denoted as $F^l_P$ and $F^l_O$, respectively. Then, we have
\begin{equation}
  F^{l}_{P} = W^{l}_{P} \circledast  F^{l-1}_{P}\,,
\end{equation}
in which $\circledast$ is the convolution operator. In this few-sample compression problem, our goal is to maximize the accuracy of the pruned model $\mathcal{M}_{P}$ using only $\mathcal{D}_{few}$.

\subsection{The need to abolish the layer-wise scheme}

Compressing with few samples is a resource constrained problem, only few training images $\mathcal{D}_{few}$ and a well trained model $\mathcal{M}_{O}$ are available to help recover the accuracy of $\mathcal{M}_{P}$. Previous works~\cite{Li_2020_CVPR,Bai_Wu_King_Lyu_2020} followed the training framework of FitNet~\cite{Romero_fitnets_2015}, which uses a layer-wise reconstruction manner to resume the representation ability of each layer. Specifically, the optimization of one specific layer is 
\begin{equation}
\mathop{\arg\min}\limits_{W^{l}_{P}} \  \mathcal{L} (W_{P}^{l}, W_{O}^{l}, F^{l-1}_{P}, F^{l-1}_{O})\,.
\end{equation}
The loss $\mathcal{L}$ usually measures the representative disparity of weights $W_{P}^{l}$ and $W_{O}^{l}$, and the most simple measure is 
\begin{equation}
  \mathcal{L}  = \|W_{P}^{l} \circledast F^{l-1}_{P} - W_{O}^{l} \circledast F^{l-1}_{O} \|^{2}_{F}\,,
\end{equation}
in which $\|X\|_F$ is the Frobenius norm of the matrix $X$ and $\|X\|_F=\sqrt{\mathrm{tr}(X^TX)}=\sqrt{\sum_{i,j}X_{i,j}^2}$. With some channels pruned, the output dimensionality of layer $l$ will change and cause a dimension mismatch issue. To handle this problem, FSKD~\cite{Li_2020_CVPR} tried to ignore the pruned features and only reconstructed the responses of the preserved channels. CD~\cite{Bai_Wu_King_Lyu_2020} pruned the convolution layers within residual blocks only, which can keep the output dimensionality of residual blocks unchanged, and then computed the loss between the features of the pruned and the original model after residual blocks. Recent pruning methods~\cite{luo_neural_2020,liu_fisherpruning_2021} improved the way to prune residual connections, which may influence many coupled convolution layers across different residual blocks. \cite{luo_neural_2020,liu_fisherpruning_2021} showed that it is \emph{highly disadvantageous} in terms of both acceleration and accuracy to only prune inside the residual connections. Hence, existing few-sample compressions methods lead to inferior pruned network structures (\cf Fig. \ref{fig1}).

Apart from the dimensionality issue, this layer-wise reconstruction framework suffers from error accumulation, which is a severe problem that gradually accumulates estimation errors along the forward path. The reconstruction in a layer tries to reduce the dissimilarity of two outputs, which will never become zero in practice, and the dissimilarity will accumulate and enlarge itself, due to not only the two models' different capacities but also the adverse impact of limited training data. To reduce error accumulation, CD~\cite{Bai_Wu_King_Lyu_2020} tried to prune conv. layers within residual blocks and only in shallow layers, while deeper conv. layers were kept unchanged (\cf Fig. \ref{fig1}).

It is also shown to be effective in model pruning to drop an entire residual block (or several blocks), which obviously breaks the one-to-one correspondence between layers of the pruned and the original model. These recent and effective pruning approaches render existing layer-wise few-shot compression scheme unusable---How can we reduce the estimation error when we do not know what to estimate? When this scheme is applicable, it is still confined by only pruning convolution layers within residual blocks.

Instead, we argue that we need a new few-sample pruning paradigm that is both general (\ie, applicable to all sorts of pruning schemes and network architectures) and highly accurate.

\subsection{Mimicking then Replacing}

Our solution is conceptually very simple: Mimicking then Replacing (MiR). As the name suggests, MiR first urges the pruned model to output the same feature representations as those of the original model for the same image. This part is different from normal representation learning, because it does not need any head (such as a classification or detection head). We only need to pickup one layer in front of such a head, and then \emph{mimic features at that layer}. In other words, this step is unsupervised and we do not need any labels.

In the second (`replacing') step, after we get a smaller student which produces nearly the same activations as those of the bigger teacher, we then replace the teacher's backbone with the smaller student but \emph{keep the head unchanged} to obtain the final compressed model. This step is obviously unsupervised, too. It is also easy to deduce that we make \emph{no} assumption on the network's structure (\ie, it is widely applicable). That is, MiR is not only simple and unsupervised, but also general.

So now the key question is: what to mimic? In knowledge distillation, researchers tried to find good supervision signals from the teacher model, and the most popular way is to use the softmax outputs (soft logits). But Wang \etal argued that the softmax outputs contain less information, and that the teacher's features in the penultimate layer (after the final pooling layer and before the classification or detection head) is a better supervision~\cite{wang_mimicking_2021}. They mimic these features directly, and focused more on feature directions and gave freedom to feature magnitudes. Therefore, they proposed an LSH loss along with the mean squared loss ($\ell_2$ loss) and the cross-entropy loss ($\mathcal{L}_{CE}$). The LSH loss is used for relaxing the constraints to magnitude. Hence, the total loss in LSHKD~\cite{wang_mimicking_2021} is 
\begin{equation}
  \mathcal{L}_{total} = \mathcal{L}_{CE} + \beta (\mathcal{L}_{mse} + \mathcal{L}_{lsh}) \,,
\end{equation}
where $\beta$ is a loss balancing hyperparameter. We follow the feature mimicking idea of LSHKD~\cite{wang_mimicking_2021}, but allows more freedom in choosing the layer for feature mimicking (\ie, mimicking features in one of the layers, but not necessarily the penultimate layer).

Because in the replacing step, the classification head remains intact, we want the student's features to be \emph{exactly} the same as those of the teacher's. Hence, we only use the $\mathcal{L}_{mse}$ term, with a nice byproduct being the $\beta$ hyperparameter eliminated. No extra hyperparameters have been introduced in our MiR.

\subsection{Mimicking features before pooling}

As will be shown, MiR, by mimicking features in the penultimate layer, has at least 2\% gain over layer-wise reconstruction methods. But, the penultimate layer's features are obtained after a pooling layer (mostly global average or max pooling). The pooling operation can filter noise in feature maps but may also filter detailed information away. Considering this, we change the mimicking target from the features after the final pooling to the features before it. The details filtered away by the pooling layer help us obtain much better results without extra computation, as our experiments will show later. Our optimization target is then
\begin{equation}
  \min \| F_{P}^L - F_{O}^L \|^2_{F}\,,
\end{equation}
where $L$ is the index of the layer whose features are being mimicked (either before or after the final pooling layer).

\section{Experiments}\label{experiments}

In this section, we verify our statements about the MiR framework through experiments. We focus on the effectiveness and generality. Hence, we experimented MiR on 1) different pruning strategies, which contains different pruning ratios and various pruning schemes; 2) different model structures (ResNet~\cite{he_deep_2016} and MobileNetV2~\cite{sandler_mobilenetv2_2018}). We report both the average top-1 and top-5 accuracy and the standard deviation with five independent trials. As for the number of samples we use, we randomly sample $K$ instances in $N$ classes ($N$ way $K$ shot, denoted as $N/K$) and also randomly sample 50/100/500 instances regardless of classes. Concretely, we tried $K=1,2,3$, and used 50/100/500 independently sampled images.

We compared our method with 1) fine-tuning with the sampled subset directly with the cross-entropy loss (denoted as BP); 2) training the pruned model with both hard targets (labels) and soft targets (softmax outputs of the original model), and denoted as KD \cite{hinton_kd_2015}; 3) FSKD \cite{Li_2020_CVPR} and 4) CD \cite{Bai_Wu_King_Lyu_2020}. The results of these compared methods were also reported after five independent trials. We implemented BP and KD by ourselves, implemented FSKD according to their official code snippets, and run the official CD codes to obtain results (cf. the appendix for more details).

For our method, we optimized with SGD, and the learning rate, weight decay, and momentum were 0.02, 1e-4, and 0.9, respectively. We decreased the learning rate by a factor of 10 per 40\% iterations. $\text{MiR}_{after}$ and $\text{MiR}_{before}$ represent mimicking features after and before the final pooling layer, respectively. On ILSVRC-2012 \cite{russakovsky_imagenet_2015}, we fine-tuned models for 2000 iterations, and the batch size is 64 (the same settings as CD). Moreover, we also fine-tuned the baseline methods (BP and KD) for 2000 iterations.

We are not studying the optimal way to prune filters, so we pruned models by the simple $\ell_1$-norm~\cite{he_channel_2017, zhu2017prune}, and kept the same keep ratio for every layer. Based on the same pruned models, we compare our method with others.

\subsection{Effectiveness} 

To show the effectiveness, we experimented with ResNet-34 \cite{he_deep_2016} on ILSVRC-2012~\cite{russakovsky_imagenet_2015}, using the ResNet-34 model from the PyTorch official site, which has 3.7G FLOPs, 21.8M parameters, and 73.3\%/91.4\% top-1/top-3.7G FLOPs and 21.8M parameters.

In Table~\ref{tab:normal_pruning}, we pruned all convolution layers within residual blocks (`Prune-C Normal' in Table~\ref{tab:prne_setting}). As described in Table~\ref{tab:prne_setting}, models with 31.3\% FLOPs and 30.3\% parameters pruned are used here. As shown in Table~\ref{tab:normal_pruning}, our MiR outperforms other layer-wise reconstruction methods (FSKD and CD) and softmax outputs optimization methods (BP and KD) by large margins consistently, especially when the number of data decreases. BP and KD methods are highly overfitting, whose accuracy on the training set were nearly 100\%, but the accuracy on the validation set were very low. The layer-wise reconstruction methods behaved better than BP and KD, and CD behaved much better than FSKD. When we used the features after pooling in MiR, we already significantly outperformed other methods. Furthermore, when we used the features before pooling as proposed in this paper, we obtained extra 1-3\% gains. This seemingly very simple change leads to sizeable improvements in accuracy consistently in all our experiments. 

Although previous works mostly report only the top-5 accuracy, Table~\ref{tab:normal_pruning} reveals that the top-1 accuracy is a more powerful evaluation metric than top-5.

\begin{table}
  \small
  \setlength\tabcolsep{5pt}
  \caption{Details of three pruning settings of different pruning schemes. The original ResNet-34 model for ILSVRC-2012 has 3.7G FLOPs and 21.8M parameters. $\downarrow$ means the percentage of reduction.}
  \centering
  \begin{tabular}{ccccc}
	\toprule									
		&        		& 	keep ratio	& 	FLOPs $\downarrow$	& 	Params $\downarrow$ 	\\
	\midrule													
	\multirow{3}{*}{Prune-A} 	&        	CD	& 	$0.70$	& 	$14.0\%$ 	& 	$7.3\%$ 	\\
		&        	Normal	& 	$0.85$	& 	$13.7\%$ 	& 	$15.0\%$ 	\\
		&        	Residual	& 	$0.93$	& 	$13.4\%$	& 	$8.8\%$	\\
	\midrule									
	\multirow{3}{*}{Prune-B} 	&        	CD	& 	$0.50$	& 	$23.7\%$	& 	$11.7\%$ 	\\
		&        	Normal	& 	$0.76$	& 	$23.8\%$	& 	$23.3\%$	\\
		&        	Residual	& 	$0.85$	& 	$23.8\%$	& 	$20.6\%$	\\
	\midrule									
	\multirow{3}{*}{Prune-C} 	&        	CD	& 	$0.30$	& 	$33.3\%$	& 	$16.1\%$	\\
		&        	Normal	& 	$0.68$	& 	$31.3\%$	& 	$30.3\%$	\\
		&        	Residual	& 	$0.80$	& 	$33.5\%$	& 	$23.5\%$	\\
	\bottomrule									
  \end{tabular}
\label{tab:prne_setting}
\end{table}

\begin{table*}
\caption{Mean and standard deviation of top-1 accuracy (\%) on ILSVRC-2012. We compare the results under the same FLOPs reduction, but with different pruning schemes. 500 randomly sampled images were used for training the pruned ResNet-34 models.}
	\centering
	\small
	\begin{tabular}{ccccccccc}
		\toprule
		\multirow{2}{*}{Methods} 
		& \multicolumn{2}{c}{Prune-A (14\% FLOPs $\downarrow$)}
		& \multicolumn{2}{c}{Prune-B (24\% FLOPs $\downarrow$)}
		& \multicolumn{2}{c}{Prune-C (33\% FLOPs $\downarrow$)} \\
		& \multicolumn{1}{l}{CD-style} & \multicolumn{1}{l}{Normal} 
		& \multicolumn{1}{l}{CD-style} & \multicolumn{1}{l}{Normal} 
		& \multicolumn{1}{l}{CD-style} & \multicolumn{1}{l}{Normal} \\
		\midrule																											
		BP	&	$65.02_{\pm 0.30}$	&	$63.43_{\pm 0.20}$	&	$58.94_{\pm 0.36}$	&	$57.67_{\pm 0.27}$	&	$47.54_{\pm 0.41}$	&	$51.90_{\pm 0.34}$	\\
		KD	&	$67.22_{\pm 0.18}$	&	$65.75_{\pm 0.13}$	&	$61.01_{\pm 0.24}$	&	$60.23_{\pm 0.16}$	&	$49.34_{\pm 0.25}$	&	$54.76_{\pm 0.19}$	\\
		FSKD	&	$69.59_{\pm 0.09}$	&	$68.75_{\pm 0.08}$	&	$62.56_{\pm 0.13}$	&	$63.72_{\pm 0.13}$	&	$33.19_{\pm 0.60}$	&	$57.65_{\pm 0.18}$	\\
		CD	&	$71.12_{\pm0.06}$	&	$69.94_{\pm0.07}$	&	$68.17_{\pm0.07}$	&	$67.13_{\pm0.06}$	&	$59.65_{\pm0.12}$	&	$63.70_{\pm0.18}$	\\
		\midrule													
		$\text{MiR}_{after}$	&	$71.64_{\pm0.08}$	&	$70.60_{\pm0.06}$	&	$69.75_{\pm0.10}$	&	$68.30_{\pm0.06}$	&	$66.41_{\pm0.06}$	&	$65.37_{\pm0.03}$	\\
		$\text{MiR}_{before}$	&	$71.95_{\pm0.07}$	&	$71.10_{\pm0.06}$	&	$70.53_{\pm0.10}$	&	$69.18_{\pm0.05}$	&	$68.14_{\pm0.04}$	&	$66.98_{\pm0.09}$	\\
		\bottomrule
	\end{tabular}
	\label{tab:cd_pruning}
\end{table*}

\begin{table*}
	\footnotesize
	\setlength\tabcolsep{1.325pt}
	\caption{Mean and standard deviation of top-1/top-5 accuracy (\%) on ILSVRC-2012. We pruned ResNet-34 using `Prune-C Residual' (\cf Table~\ref{tab:prne_setting}).}
	\centering
	\begin{tabular}{ccccccc}
		\toprule
		Methods & 50 & 100 & 500 & 1000/1 & 1000/2 & 1000/3 \\
		\midrule
		BP	&	$24.2_{\pm0.92}	/	{52.7_{\pm1.36}}$	&	$27.6_{\pm0.41}	/	{56.7_{\pm0.62}}$	&	$42.9_{\pm0.28}	/	{70.5_{\pm0.27}}$	&	$51.2_{\pm0.32}	/	{76.5_{\pm0.16}}$	&	$54.6_{\pm0.26}	/	{79.0_{\pm0.10}}$	&	$56.0_{\pm0.17}	/	{80.1_{\pm0.10}}$	\\
		KD	&	$30.1_{\pm0.69}	/	{57.7_{\pm1.10}}$	&	$33.1_{\pm0.43}	/	{61.0_{\pm0.53}}$	&	$45.7_{\pm0.26}	/	{72.2_{\pm0.25}}$	&	$50.5_{\pm0.29}	/	{75.9_{\pm0.23}}$	&	$52.3_{\pm0.14}	/	{77.3_{\pm0.08}}$	&	$52.7_{\pm0.11}	/	{77.6_{\pm0.09}}$	\\
		FSKD	&	$31.1_{\pm0.90}	/	{56.5_{\pm1.10}}$	&	$36.6_{\pm0.44}	/	{63.1_{\pm0.46}}$	&	$42.8_{\pm0.49}	/	{69.1_{\pm0.58}}$	&	$44.9_{\pm0.20}	/	{70.5_{\pm0.29}}$	&	$45.4_{\pm0.23}	/	{70.9_{\pm0.33}}$	&	$45.6_{\pm0.14}	/	{71.0_{\pm0.12}}$	\\
		$\text{MiR}_{after}$	&	$53.4_{\pm0.40}	/	{78.6_{\pm0.37}}$	&	$56.6_{\pm0.43}	/	{81.2_{\pm0.30}}$	&	$62.4_{\pm0.14}	/	{85.1_{\pm0.11}}$	&	$64.3_{\pm0.07}	/	{86.2_{\pm0.06}}$	&	$65.3_{\pm0.10}	/	{86.8_{\pm0.03}}$	&	$65.8_{\pm0.05}	/	{87.2_{\pm0.03}}$	\\
		$\text{MiR}_{before}$	&	$59.9_{\pm0.30}	/	{83.2_{\pm0.31}}$	&	$62.1_{\pm0.22}	/	{84.8_{\pm0.18}}$	&	$65.4_{\pm0.07}	/	{87.0_{\pm0.03}}$	&	$66.6_{\pm0.05}	/	{87.7_{\pm0.04}}$	&	$67.2_{\pm0.05}	/	{88.2_{\pm0.05}}$	&	$67.5_{\pm0.04}	/	{88.3_{\pm0.05}}$	\\
		\bottomrule
	\end{tabular}
	\label{tab:residual_pruning}
\end{table*}

\subsection{Different pruning ratios and schemes}

As aforementioned, a good framework should be general. In this subsection, we apply our MiR framework on pruned models with different FLOPs and different pruning schemes to show the generality of the MiR framework.

Precisely, by `pruning schemes' we refer to the way that is used to trim the model. Three pruning schemes are used in this paper, which are visualized in Fig. \ref{fig1} and their statistics are reported in Table~\ref{tab:prne_setting}. As shown in Fig. \ref{fig1}, the `CD-style' pruning scheme only prunes some shallow layers within residual blocks~\cite{Bai_Wu_King_Lyu_2020}. This setting makes sure the reduction of representation ability only occurs in shallow layers, and the accumulated error through shallow layers can be compensated by those unpruned deeper layers. In contrast, the `Normal'-style pruning scheme is the way most researchers use, which prunes all layers within residual blocks. Furthermore, recent methods~\cite{luo_neural_2020, liu_fisherpruning_2021} proposed to prune the residual connection (or coupled convolution), which is believed to be a more reasonable way to obtain compact models. With `Residual'-style pruning, coupled blocks will be affected simultaneously (as shown in Fig. \ref{fig1}), and the output dimensionality of each block will change, too. 

We use Prune-A to represent different models with the same FLOPs pruned, and so do Prune-B and Prune-C. As reported in Table~\ref{tab:prne_setting}, Prune-A/B/C represent models with around 14\%, 24\%, and 33\% FLOPs reduction, in which keep ratio means the ratio of output channels kept in layers after pruning. It is worth mentioning that `Prune-B CD' is the same as the setting Res-50\% used in CD~\cite{Bai_Wu_King_Lyu_2020}. As we can see, these three pruning schemes have different model size reduction under the same FLOPs, which is because in ResNet, blocks in shallow layers have nearly the same FLOPs but much fewer parameters than deeper layers. We will soon see that these three pruning schemes have pros and cons in different aspects. 

To show the performance under different pruning schemes and different pruning ratios, we report the results of `CD' and `Normal' style with 500 samples used. As reported in Table~\ref{tab:cd_pruning}, we have the following findings according to these results:

\squishlist
  \item All methods behave well when pruning a small number of FLOPs, and with more FLOPs pruned, the fine-tuned accuracy drops quickly. Our MiR (both $\text{MIR}_{after}$ and $\text{MIR}_{before}$) outperforms others with a large margin. Changing the mimicked features from after pooling to before pooling has a significant increase in accuracy, and the gap expands when more FLOPs are pruned. It is worth noting that MiR is in general more stable (\ie, smaller standard deviations) and is unsupervised. In contrast, BP, KD and CD need to use image labels in $\mathcal{D}_{few}$.
  \item In Prune-A and Prune-B, the fine-tuning results of the `CD-style' scheme are higher compared with the `Normal' scheme. There are two possible reasons. The first is that '`CD-style' pruning preserves more parameters than `Normal' under the same FLOPs. The second is that `CD-style' pruning preserves more channels in deeper layers, and it is easier to recover the pruned models with these preserved deep layers.
  \item As Table~\ref{tab:prne_setting} shows, `Prune-C CD-style' trims 70\% channels in shallow layers. The pruned network is of an hourglass shape~\cite{luo_neural_2020} and most information is lost in these layers. Hence, the CD method becomes unstable in this case. Since we optimize parameters in the backbone globally and the information can flow through the residual connections, our MiR method is still stable and accurate. 
\squishend

\subsection{Pruning residual connections}

The CD method needs to keep the dimensionality of the compressed model's feature maps the same as that of the original model. Because the residual connection pruning scheme changes the dimensionality, the CD method is not usable in the residual pruning scheme. Therefore, we compare MiR with BP, KD, and FSKD in Table~\ref{tab:residual_pruning} for the `Prune-C Residual' scheme. When comparing these results with the `Prune-C 500 samples' results in Table~\ref{tab:prne_setting}, we find the `Residual' scheme is harder than the `CD-style' and `Normal' pruning schemes. Now the layer-wise reconstruction method FSKD is highly ineffective in the residual pruning scheme, which even has lower accuracy than BP and KD, but FSKD behaved better than BP and KD in the `Normal' pruning scheme. We also note that the `CD-style' pruning not only has fewer FLOPs reduction (14\% vs. 33\% of `Residual'), the network's hourglass shape also makes it slower than `Residual' even when the FLOPs are the same~\cite{luo_neural_2020}. Hence, the `Residual' pruning scheme is more useful in practice than `CD-style' and `Normal'. In the `Residual' scheme, our MiR still has the highest accuracy.

\subsection{Results on MobileNetV2}

To further validate the generality of our MiR framework, we implemented MiR on MobileNetV2~\cite{sandler_mobilenetv2_2018}, which is widely used in edge devices and has a different structure as the ResNet series. Instead of expanding MobileNetV2 then pruning it (widely adopted in pruning methods), we directly prune official MobileNetV2-$1.0\times$ with different FLOPs, \ie Prune-D and Prune-E, using the `Normal' scheme. Prune-D and Prune-E prune 25\% and 15\% channels in each layer, respectively. So Prune-D prunes 21.6\% FLOPs and 12.9\% parameters, Prune-E prunes 13.3\% FLOPs and 7.7\% parameters. We compare BP and KD methods with MiR using 500 samples or 1000-way-3-shot samples, with results in Table~\ref{tab:mobilenet_pruning}. The original MobileNetV2 in ILSVRC-2012 from official PyTorch website has 71.9\% top-1 and 90.3\% top-5. $\text{MIR}_{after}$ and $\text{MIR}_{before}$ work well in MobileNetV2.

\begin{table}
	\footnotesize
	\setlength\tabcolsep{2pt}
	\centering
	\caption{Mean and standard deviation of top-1/top-5 accuracy (\%) on ILSVRC-2012. We pruned MobileNetV2 with the `Normal' pruning scheme and pruned into different FLOPs.}
	\begin{tabular}{cccc}
		\toprule										
		&	Methods	&			500	&	1000/3			\\
		\midrule
		&	\footnotesize{MobileNetV2}	&	$71.9$	/	$90.3$	&	$71.9$	/	$90.3$	\\
		\midrule\multirow{3}{*}{Prune-D} 										
		&	BP	&	$45.0_{\pm 0.34}	/	{71.8_{\pm 0.38}}$	&	$59.1_{\pm 0.22}	/	{82.0_{\pm 0.14}}$	\\
		&	KD	&	$48.4_{\pm 0.34}	/	{73.9_{\pm 0.32}}$	&	$57.5_{\pm 0.21}	/	{80.8_{\pm 0.08}}$	\\
		&	$\text{MiR}_{after}$	&	$66.0_{\pm 0.11}	/	{87.0_{\pm 0.09}}$	&	$67.1_{\pm 0.11}	/	{87.8_{\pm 0.05}}$	\\
		&	$\text{MiR}_{before}$	&	$67.6_{\pm 0.05}	/	{87.9_{\pm 0.04}}$	&	$68.3_{\pm 0.05}	/	{88.4_{\pm 0.05}}$	\\
		\midrule\multirow{3}{*}{Prune-E}										
		&	BP	&	$55.5_{\pm 0.16}	/	{80.3_{\pm 0.26}}$	&	$64.4_{\pm 0.15}	/	{85.7_{\pm 0.08}}$	\\
		&	KD	&	$59.1_{\pm 0.17}	/	{82.5_{\pm 0.15}}$	&	$64.5_{\pm 0.10}	/	{85.7_{\pm 0.05}}$	\\
		&	$\text{MiR}_{after}$	&	$68.9_{\pm 0.03}	/	{88.8_{\pm 0.05}}$	&	$69.3_{\pm 0.06}	/	{89.1_{\pm 0.03}}$	\\
		&	$\text{MiR}_{before}$	&	$69.7_{\pm 0.04}	/	{89.2_{\pm 0.03}}$	&	$69.9_{\pm 0.02}	/	{89.4_{\pm 0.03}}$	\\
		\bottomrule	
	\end{tabular}
	\label{tab:mobilenet_pruning}
\end{table}

\section{Further Analyses}

In this section, we further analyze the impacts of hyperparameters, number of training iterations, and training set sizes. We also report the results with different loss function, and examine the limitations of MiR.

\subsection{Hyperparameters}\label{ablation:hyper}

As aforementioned, MiR has no extra hyperparameters, except those already in the optimizer. In our experiments, we only changed the initial learning rate in SGD. To explore the influence of the initial learning rate, we tried different values in both $\text{MiR}_{after}$ and $\text{MiR}_{before}$. Experiments indicate that 0.02 is a good initial learning rate. Too large or too small values are harmful (\cf results in Table~\ref{tbl:lr}). We trained models of `Prune-B CD-style' (which is also Res-50\% in CD~\cite{Bai_Wu_King_Lyu_2020}) with 500 randomly sampled images, and we report the mean top-1 accuracy of five independent trials.

\begin{table}
	\small
	\caption{Average top-1 accuracy on ILSVRC-2012 under different initial learning rates. Models are pruned using `Prune-B CD-style' and 500 samples.}
	\label{tbl:lr}
	\centering
	\setlength\tabcolsep{3pt}
	\renewcommand{\arraystretch}{1.1}
	\begin{tabular}{ccccccccc}
		\toprule
		&	0.1	&	0.05	&	0.02	&	0.01	&	0.005	&	0.002	&	0.001	\\
		\midrule
		$\text{MiR}_{after}$	&	$68.51$	&	$70.02$	&	$\mathbf{70.53}$	&	$70.50$	&	$70.23$	&	$69.54$	&	$68.89$	\\
		$\text{MiR}_{before}$	&	$68.77$	&	$69.73$	&	$\mathbf{69.74}$	&	$69.42$	&	$68.93$	&	$68.06$	&	$67.25$	\\
		\bottomrule
	\end{tabular}
\end{table}

\subsection{Training time and training set size}

\begin{figure}
  \centering   
  \includegraphics[width=0.85\linewidth]{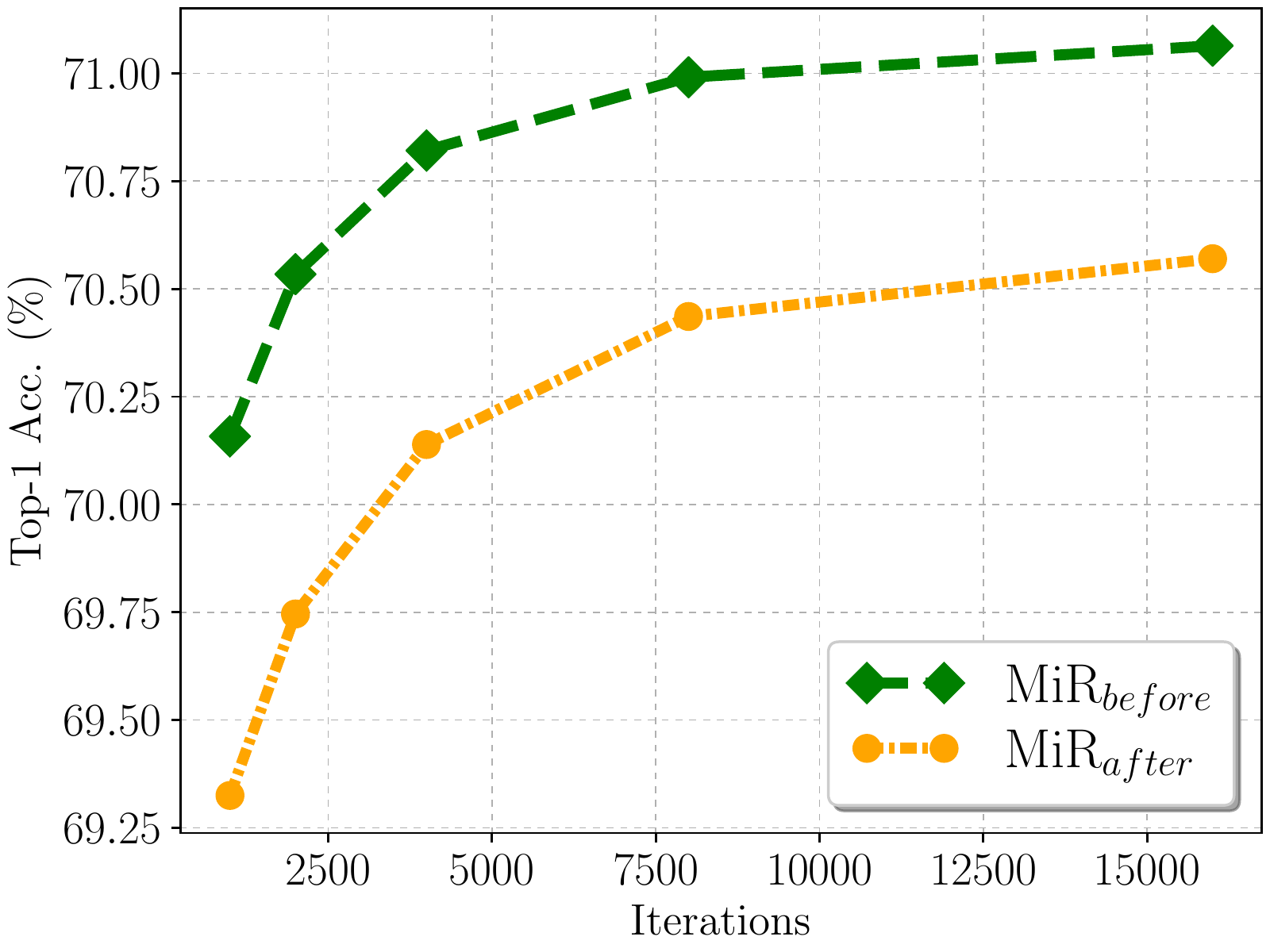}
    \caption{Average top-1 accuracy with different numbers of training iterations. Best viewed in color.}
    \label{fig:sub1}
\end{figure}

We further explore the influence of training time (number of training iterations) and training set size (number of training samples). First of all, we fine-tuned the pruned models with 1k, 2k, 4k, 8k, 16k iterations with 500 samples, using the `Prune-B CD-style' setting the same as in Sec.~\ref{ablation:hyper}. As shown in Fig.~\ref{fig:sub1}, the accuracy increases when more iterations are used on both $\text{MiR}_{after}$ and $\text{MiR}_{before}$, but the speed of increase gradually diminishes. One possible explanation for the accuracy boost is: Since we used the standard data augmentation (random flip and random crop), the randomness in data augmentation brings in similar but different representations (\ie, features or activation maps) of each image, which provides more information for the pruned model to mimic.

\begin{figure}
  \centering   
  \includegraphics[width=0.85\linewidth]{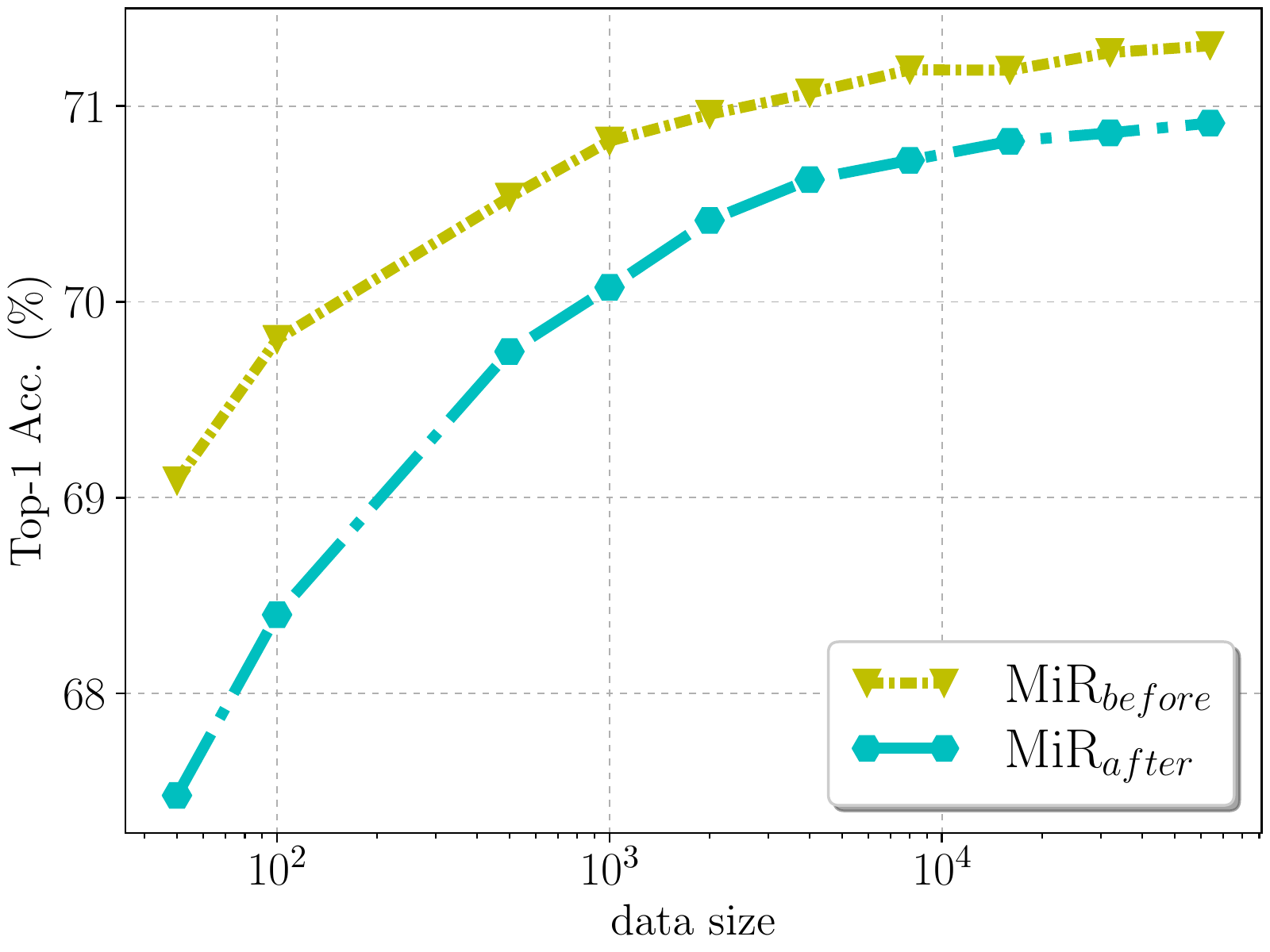}
    \caption{Average top-1 accuracy with different number of training samples. Best viewed in color.}
    \label{fig:sub2}
\end{figure}

Next, we also analyze the impact of the number of training data. We used 500, 1k, 2k, 4k, 8k, 16k, 32k randomly sampled images for training, and report the mean accuracy with 2000 iterations fine-tuning. According to the results shown in Fig.~\ref{fig:sub2}, there is no doubt that using more training images lead to better accuracy, especially when we start from a tiny training set. Moreover, we fine-tuned 16k iterations with 10k randomly sampled images. We achieve 71.8\% and 90.6\% top-1 and top-5 accuracy, respectively. Compared with the original ResNet-34 (top-1/top-5 is 73.3\%/91.4\%), we compressed a model by 24\% FLOPs reduction and only 0.8\% top-5 accuracy drop, which took less than one hour in a single 32G V100 GPU with less than 1\% of the original training set.

\begin{table}
  \caption{Results of different loss functions on ILSVRC-2012. We trained models pruned by `Prune-B CD-style' and 500 samples. Mean and std. of Top-1 and Top-5 acc. are reported.}
  \centering
  \small
  \begin{tabular}{cccccc}
	\toprule					
	MSE	 & $\ell_1$-norm	& sim loss & LSH & top-1/top-5	\\
	\midrule					
	\cmark  &         &           &            &	$69.75/89.29$		\\
			&  \cmark &           &            &	$69.88/89.33$		\\
			&         &  \cmark   &            &	$69.07/88.83$		\\
			&         &           &   \cmark   &	$66.89/87.59$		\\
	\cmark  &  \cmark &           &            &	$69.89/89.40$		\\
	\cmark  &         &  \cmark   &            &	$69.82/89.32$		\\
	\cmark  &         &           &   \cmark   &	$69.26/89.07$		\\			
	\bottomrule								
  \end{tabular}
  \label{ablation:losses}
\end{table}

\subsection{Comparison with other loss functions}

In our Mimicking then Replacing framework, it is easy to extend it by changing or adding another loss function. In this part, we try to mimic features with some other loss functions, and compare their performance in our MiR framework. Because we aim at mimicking the responses rather than performing representation learning, we compare the $\ell_2$ loss (using the features after pooling) with 1) the $\ell_1$ loss; 2) maximizing the cosine similarity (`sim loss' in Table \ref{ablation:losses}); 3) the LSH loss in~\cite{wang_mimicking_2021}. 

As shown in Table~\ref{ablation:losses}, both MSE ($\ell_2$) and $\ell_1$ based loss functions fit well in mimicking features for few-sample compression. Because the LSH loss relaxes the constraints to feature magnitudes, it is not as effective in this few-sample compression task.

\subsection{Freeze backbone and train classifier}

As the results in Sec.~\ref{experiments} shows, it is vulnerable to overfitting if we fine-tune the whole network (both backbone and head). In our Mimicking then Replacing framework, we only train the layers before the classifier head (\ie, the backbone) and directly use the head from the original model. We need to decide whether the replacing operation is a good choice. Therefore, we compare our MiR results with 1) mimicking the features first and then freeze the backbone to learn the classifier; and, 2) freeze the backbone without fine-tuning, and then tuning the classifier. 

The results are in Table~\ref{ablation:tuning}. When using a larger learning rate to tune the classifier while freezing the backbone, the accuracy will drop. When using a small enough learning rate, the weights in the classifier are almost not updated. These results show that keeping the classifier head unchanged is better than tuning the classifier.

\begin{table}
	\caption{The average top-1/top-5 accuracy when tuning classifier under freezing trained or untrained backbone. Without tuning classifier, we have 69.75/89.29 when training backbone using MiR.}
	\label{ablation:tuning}
	\centering
	\small
	\begin{tabular}{ccc}
		\toprule
		\multirow{2}{*}{LR} &	freeze	&	freeze	\\
		&	trained backbone	& untrained backbone	\\
		\midrule
		0.1	&	$66.44/87.96$	&	$45.43/73.24$ \\
		0.05 & $69.96/88.97$ & $50.52/76.45$ \\
		0.01 & $69.52/89.22$ & $52.37/77.53$ \\
		0.005 & $69.59/89.21$ & $52.08/77.29$ \\
		0.001 & $69.60/89.24$ & $51.33/76.46$ \\
		\bottomrule
	\end{tabular}
\end{table}

\subsection{Limitations}

In our Mimicking then Replacing framework, we aim at obtaining a compact backbone that behaves almost the same as the original one, which means we can only get a model with the same or weaker representation ability than the original model. The accuracy of the pruned model is bounded from above by the accuracy of the original model. 

And there are two potential directions to extend our MiR framework. The first one is to augment the input data, which may provide more information for feature mimicking. The other one is to add loss functions, which also means adding more supervision signals.

\section{Conclusion}

In this paper, we proposed a new framework, Mimicking then Replacing, for few-sample compression, which is not only simple and unsupervised, but also general and highly accurate. Unlike previous layer-wise reconstruction methods, we directly urged the student to mimic the teacher's features around the penultimate layer, which made no assumption on the network's structure and made MiR general to use. We followed the feature mimicking idea of LSHKD~\cite{wang_mimicking_2021}, but we further mimicked features before the final pooling layer, leading to significant improvements without introducing extra computations.

{\small
\bibliographystyle{ieee_fullname}
\bibliography{cvpr.bib}

\begin{thebibliography}{10}\itemsep=-1pt

\bibitem{Bai_Wu_King_Lyu_2020}
Haoli Bai, Jiaxiang Wu, Irwin King, and Michael Lyu.
\newblock Few shot network compression via cross distillation.
\newblock In {\em Proceedings of the AAAI Conference on Artificial
  Intelligence}, pages 3203--3210, 2020.

\bibitem{Chen_2019_ICCV}
Hanting Chen, Yunhe Wang, Chang Xu, Zhaohui Yang, Chuanjian Liu, Boxin Shi,
  Chunjing Xu, Chao Xu, and Qi Tian.
\newblock Data-free learning of student networks.
\newblock In {\em Proceedings of the IEEE International Conference on Computer
  Vision (ICCV)}, pages 3514--3522, 2019.

\bibitem{chen_network_2020}
Zhengsu Chen, Jianwei Niu, Lingxi Xie, Xuefeng Liu, Longhui Wei, and Qi Tian.
\newblock Network adjustment: Channel search guided by {FLOPs} utilization
  ratio.
\newblock In {\em Proceedings of the IEEE Conference on Computer Vision and
  Pattern Recognition (CVPR)}, pages 10658--10667, 2020.

\bibitem{choudhary2020comprehensive}
Tejalal Choudhary, Vipul Mishra, Anurag Goswami, and Jagannathan Sarangapani.
\newblock A comprehensive survey on model compression and acceleration.
\newblock {\em Artificial Intelligence Review}, 53(7):5113--5155, 2020.

\bibitem{deng_compression_survey_2020}
Lei Deng, Guoqi Li, Song Han, Luping Shi, and Yuan Xie.
\newblock Model compression and hardware acceleration for neural networks: A
  comprehensive survey.
\newblock {\em Proceedings of the IEEE}, 108(4):485--532, 2020.

\bibitem{dubey2018coreset}
Abhimanyu Dubey, Moitreya Chatterjee, and Narendra Ahuja.
\newblock Coreset-based neural network compression.
\newblock In {\em Proceedings of the European Conference on Computer Vision
  (ECCV)}, volume 11220 of {\em LNCS}, pages 454--470. Springer, 2018.

\bibitem{gholami2021survey}
Amir Gholami, Sehoon Kim, Zhen Dong, Zhewei Yao, Michael~W Mahoney, and Kurt
  Keutzer.
\newblock A survey of quantization methods for efficient neural network
  inference.
\newblock {\em {arXiv} preprint {arXiv}:2103.13630}, 2021.

\bibitem{guo_dmcp_2020}
Shaopeng Guo, Yujie Wang, Quanquan Li, and Junjie Yan.
\newblock {DMCP}: Differentiable {Markov} channel pruning for neural networks.
\newblock In {\em Proceedings of the IEEE Conference on Computer Vision and
  Pattern Recognition (CVPR)}, pages 1539--1547, 2020.

\bibitem{guo2016dynamic}
Yiwen Guo, Anbang Yao, and Yurong Chen.
\newblock Dynamic network surgery for efficient dnns.
\newblock In {\em Advances in Neural Information Processing Systems 29
  (NeurIPS)}, 2016.

\bibitem{han2015learning}
Song Han, Jeff Pool, John Tran, and William~J Dally.
\newblock Learning both weights and connections for efficient neural networks.
\newblock In {\em Advances in Neural Information Processing Systems 28
  (NeurIPS)}, 2015.

\bibitem{Haroush_2020_CVPR}
Matan Haroush, Itay Hubara, Elad Hoffer, and Daniel Soudry.
\newblock The knowledge within: Methods for data-free model compression.
\newblock In {\em Proceedings of the IEEE Conference on Computer Vision and
  Pattern Recognition (CVPR)}, pages 8494--8502, 2020.

\bibitem{he_deep_2016}
Kaiming He, Xiangyu Zhang, Shaoqing Ren, and Jian Sun.
\newblock Deep residual learning for image recognition.
\newblock In {\em Proceedings of the IEEE Conference on Computer Vision and
  Pattern Recognition (CVPR)}, pages 770--778, 2016.

\bibitem{he2018soft}
Yang He, Guoliang Kang, Xuanyi Dong, Yanwei Fu, and Yi Yang.
\newblock Soft filter pruning for accelerating deep convolutional neural
  networks.
\newblock In {\em Proceedings of the International Joint Conference on
  Artificial Intelligence ({IJCAI})}, 2018.

\bibitem{he_channel_2017}
Yihui He, Xiangyu Zhang, and Jian Sun.
\newblock Channel pruning for accelerating very deep neural networks.
\newblock In {\em Proceedings of the IEEE International Conference on Computer
  Vision (ICCV)}, pages 1389--1397, 2017.

\bibitem{hinton_kd_2015}
Geoffrey Hinton, Oriol Vinyals, and Jeffrey Dean.
\newblock Distilling the knowledge in a neural network.
\newblock In {\em NIPS Deep Learning and Representation Learning Workshop},
  2015.

\bibitem{li_eagleeye_2020}
Bailin Li, Bowen Wu, Jiang Su, Guangrun Wang, and Liang Lin.
\newblock {EagleEye}: Fast sub-net evaluation for efficient neural network
  pruning.
\newblock In {\em The European Conference on Computer Vision (ECCV)}, volume
  12347 of {\em LNCS}, pages 639--654. Springer, 2020.

\bibitem{li_pruning_2017}
Hao Li, Asim Kadav, Igor Durdanovic, Hanan Samet, and Hans~Peter Graf.
\newblock Pruning filters for efficient convnets.
\newblock In {\em The International Conference on Learning Representations
  (ICLR)}, pages 1--13, 2017.

\bibitem{Li_2020_CVPR}
Tianhong Li, Jianguo Li, Zhuang Liu, and Changshui Zhang.
\newblock Few sample knowledge distillation for efficient network compression.
\newblock In {\em Proceedings of the IEEE Conference on Computer Vision and
  Pattern Recognition (CVPR)}, pages 14639--14647, 2020.

\bibitem{lin2020hrank}
Mingbao Lin, Rongrong Ji, Yan Wang, Yichen Zhang, Baochang Zhang, Yonghong
  Tian, and Ling Shao.
\newblock Hrank: Filter pruning using high-rank feature map.
\newblock In {\em Proceedings of the CVF Conference on Computer Vision and
  Pattern Recognition (CVPR)}, pages 1529--1538, 2020.

\bibitem{liu_fisherpruning_2021}
Liyang Liu, Shilong Zhang, Zhanghui Kuang, Aojun Zhou, Jing-Hao Xue, Xinjiang
  Wang, Yimin Chen, Wenming Yang, Qingmin Liao, and Wayne Zhang.
\newblock Group fisher pruning for practical network compression.
\newblock In {\em Proceedings of the 38th International Conference on Machine
  Learning (ICML)}, pages 7021--7032, 2021.

\bibitem{liu_metapruning_2019}
Zechun Liu, Haoyuan Mu, Xiangyu Zhang, Zichao Guo, Xin Yang, Kwang-Ting Cheng,
  and Jian Sun.
\newblock {MetaPruning}: Meta learning for automatic neural network channel
  pruning.
\newblock In {\em Proceedings of the IEEE International Conference on Computer
  Vision (ICCV)}, pages 3296--3305, 2019.

\bibitem{luo_neural_2020}
Jian-Hao Luo and Jianxin Wu.
\newblock Neural network pruning with residual-connections and limited-data.
\newblock In {\em Proceedings of the IEEE Conference on Computer Vision and
  Pattern Recognition (CVPR)}, pages 1458--1467, 2020.

\bibitem{luo_thinet_2017}
Jian-Hao Luo, Hao Zhang, Hong-Yu Zhou, Chen-Wei Xie, Jianxin Wu, and Weiyao
  Lin.
\newblock {ThiNet}: Pruning cnn filters for a thinner net.
\newblock {\em IEEE Transactions on Pattern Analysis and Machine Intelligence
  (TPAMI)}, 41(10):2525--2538, 2019.

\bibitem{Romero_fitnets_2015}
Adriana Romero, Nicolas Ballas, Samira~Ebrahimi Kahou, Antoine Chassang, Carlo
  Gatta, and Yoshua Bengio.
\newblock {FitNets}: Hints for thin deep nets.
\newblock In {\em The International Conference on Learning Representations
  (ICLR)}, pages 1--13, 2015.

\bibitem{russakovsky_imagenet_2015}
Olga Russakovsky, Jia Deng, Hao Su, Jonathan Krause, Sanjeev Satheesh, Sean Ma,
  Zhiheng Huang, Andrej Karpathy, Aditya Khosla, Michael Bernstein,
  Alexander~C. Berg, and Li Fei-Fei.
\newblock {ImageNet} large scale visual recognition challenge.
\newblock {\em International Journal of Computer Vision}, 115(3):211--252,
  2015.

\bibitem{sandler_mobilenetv2_2018}
Mark Sandler, Andrew Howard, Menglong Zhu, Andrey Zhmoginov, and Liang-Chieh
  Chen.
\newblock {MobileNetV2}: Inverted residuals and linear bottlenecks.
\newblock In {\em Proceedings of the IEEE Conference on Computer Vision and
  Pattern Recognition (CVPR)}, pages 4510--4520, 2018.

\bibitem{Shen_Wang_Yin_Song_Luo_Song_2021}
Chengchao Shen, Xinchao Wang, Youtan Yin, Jie Song, Sihui Luo, and Mingli Song.
\newblock Progressive network grafting for few-shot knowledge distillation.
\newblock In {\em Proceedings of the AAAI Conference on Artificial
  Intelligence}, pages 2541--2549, 2021.

\bibitem{tung2018clip}
Frederick Tung and Greg Mori.
\newblock Clip-q: Deep network compression learning by in-parallel
  pruning-quantization.
\newblock In {\em Proceedings of the IEEE Conference on Computer Vision and
  Pattern Recognition (CVPR)}, pages 7873--7882, 2018.

\bibitem{wang_mimicking_2021}
Guo-Hua Wang, Yifan Ge, and Jianxin Wu.
\newblock Distilling knowledge by mimicking features.
\newblock {\em IEEE Transactions on Pattern Analysis and Machine Intelligence
  (TPAMI)}, in press.

\bibitem{zhu2017prune}
Michael Zhu and Suyog Gupta.
\newblock To prune, or not to prune: exploring the efficacy of pruning for
  model compression.
\newblock {\em {arXiv} preprint {arXiv}:1710.01878}, 2017.

\end{thebibliography}
}

\clearpage 

\begin{appendix}

\begin{table*}
	\footnotesize
	\setlength\tabcolsep{1.325pt}
	\caption{Mean and standard deviation of top-1/top-5 accuracy (\%) on ILSVRC-2012. We used `Prune-B Normal' to prune ResNet-34 and compared different methods with different training sizes. We used 50, 100, 500 random samples, and $N$-way-$K$-shot ($N/K$ in the top row) settings. All the results were reported with five trails. \textbf{Bold} denotes the best results.}
	\centering
	\begin{tabular}{cccc|ccc}
		\toprule
		Methods	&	50			&	100			&	500			&	1000/1			&	1000/2			&	1000/3			\\
		\midrule																									
		BP	&	$48.3_{\pm1.55}	/	76.4_{\pm0.94}$	&	$49.3_{\pm0.44}	/	77.4_{\pm0.41}$	&	$57.9_{\pm0.19}	/	82.5_{\pm0.09}$	&	$62.0_{\pm0.27}	/	84.5_{\pm0.20}$	&	$63.7_{\pm0.23}	/	85.5_{\pm0.13}$	&	$64.6_{\pm0.14}	/	86.0_{\pm0.10}$	\\
		KD	&	$52.7_{\pm1.43}	/	78.8_{\pm0.99}$	&	$54.0_{\pm0.53}	/	80.1_{\pm0.47}$	&	$60.3_{\pm0.10}	/	83.8_{\pm0.09}$	&	$62.5_{\pm0.08}	/	84.9_{\pm0.08}$	&	$63.4_{\pm0.19}	/	85.4_{\pm0.08}$	&	$63.7_{\pm0.10}	/	85.6_{\pm0.07}$	\\
		FSKD	&	$55.8_{\pm0.38}	/	80.2_{\pm0.26}$	&	$59.6_{\pm0.35}	/	83.1_{\pm0.17}$	&	$63.7_{\pm0.13}	/	85.8_{\pm0.04}$	&	$64.8_{\pm0.07}	/	86.4_{\pm0.08}$	&	$65.3_{\pm0.06}	/	86.7_{\pm0.07}$	&	$65.5_{\pm0.11}	/	86.8_{\pm0.05}$	\\
		CD	&	$62.7_{\pm0.28}	/	85.1_{\pm0.19}$	&	$62.8_{\pm0.25}	/	85.2_{\pm0.15}$	&	$67.1_{\pm0.06}	/	88.0_{\pm0.05}$	&	$67.5_{\pm0.10}	/	88.2_{\pm0.04}$	&	$67.8_{\pm0.10}	/	88.4_{\pm0.06}$	&	$68.1_{\pm0.10}	/	88.5_{\pm0.05}$	\\
		\midrule																									
		$\text{MiR}_{after}$	&	$65.7_{\pm0.09}	/	87.3_{\pm0.07}$	&	$66.6_{\pm0.07}	/	87.8_{\pm0.11}$	&	$68.3_{\pm0.07}	/	88.7_{\pm0.08}$	&	$68.9_{\pm0.03}	/	89.1_{\pm0.04}$	&	$69.3_{\pm0.09}	/	89.2_{\pm0.06}$	&	$69.5_{\pm0.05}	/	89.4_{\pm0.05}$	\\
		$\text{MiR}_{before}$	&	\textbf{67.5}$_{\pm0.13}/$\textbf{88.3}$_{\pm0.06}$	&	\textbf{68.1}$_{\pm0.13}/$\textbf{88.7}$_{\pm0.07}$	&	\textbf{69.2}$_{\pm0.05}	/$\textbf{89.3}$_{\pm0.08}$	&	\textbf{69.7}$_{\pm0.06}/$\textbf{89.5}$_{\pm0.03}$	&	\textbf{69.9}$_{\pm0.04}/$\textbf{89.7}$_{\pm0.07}$	&	\textbf{70.0}$_{\pm0.07}/$\textbf{89.8}$_{\pm0.03}$	\\
		\bottomrule
	\end{tabular}
\label{app:normal_b}
\end{table*}

\begin{table*}
	\footnotesize
	\setlength\tabcolsep{1.325pt}
	\caption{Mean and standard deviation of top-1/top-5 accuracy (\%) on ILSVRC-2012. We pruned ResNet-34 using `Prune-B Residual' (\cf Table~\ref{tab:prne_setting}). \textbf{Bold} denotes the best results.}
	\centering
	\begin{tabular}{cccc|ccc}
		\toprule
		Methods & 50 & 100 & 500 & 1000/1 & 1000/2 & 1000/3 \\
		\midrule
		BP	&	$36.9_{\pm0.94}	/	66.3_{\pm1.04}$	&	$39.7_{\pm0.36}	/	69.0_{\pm0.25}$	&	$51.3_{\pm0.21}	/	77.5_{\pm0.18}$	&	$57.2_{\pm0.22}	/	81.0_{\pm0.13}$	&	$59.6_{\pm0.19}	/	82.8_{\pm0.14}$	&	$60.8_{\pm0.15}	/	83.6_{\pm0.06}$	\\
		KD	&	$42.4_{\pm0.48}	/	70.1_{\pm0.68}$	&	$44.9_{\pm0.40}	/	72.2_{\pm0.35}$	&	$54.0_{\pm0.18}	/	79.0_{\pm0.16}$	&	$57.2_{\pm0.15}	/	81.0_{\pm0.11}$	&	$58.6_{\pm0.10}	/	82.0_{\pm0.09}$	&	$58.9_{\pm0.10}	/	82.2_{\pm0.04}$	\\
		FSKD	&	$46.0_{\pm0.49}	/	72.2_{\pm0.46}$	&	$50.6_{\pm0.19}	/	76.3_{\pm0.14}$	&	$55.9_{\pm0.25}	/	80.2_{\pm0.15}$	&	$57.2_{\pm0.09}	/	81.2_{\pm0.14}$	&	$57.8_{\pm0.11}	/	81.6_{\pm0.13}$	&	$58.0_{\pm0.05}	/	81.7_{\pm0.08}$	\\
		\midrule
		$\text{MiR}_{after}$	&	$61.1_{\pm0.20}	/	84.2_{\pm0.24}$	&	$62.9_{\pm0.18}	/	85.5_{\pm0.16}$	&	$66.2_{\pm0.12}	/	87.5_{\pm0.07}$	&	$67.3_{\pm0.06}	/	88.2_{\pm0.07}$	&	$68.0_{\pm0.07}	/	88.6_{\pm0.05}$	&	$68.3_{\pm0.03}	/	88.7_{\pm0.06}$	\\
		$\text{MiR}_{before}$	&	\textbf{64.9}$_{\pm0.25}/$\textbf{86.6}$_{\pm0.21}$	&	\textbf{66.2}$_{\pm0.10}/$\textbf{87.5}$_{\pm0.12}$	&	\textbf{68.2}$_{\pm0.12}/$\textbf{88.7}$_{\pm0.05}$	&	\textbf{68.8}$_{\pm0.05} /$\textbf{89.1}$_{\pm0.06}$	&	\textbf{69.3}$_{\pm0.06}/$\textbf{89.3}$_{\pm0.03}$	&	\textbf{69.5}$_{\pm0.06}/$\textbf{89.5}$_{\pm0.03}$	\\
		\bottomrule
	\end{tabular}
	\label{app:residual_b}
	\end{table*}

\begin{table*}
	\centering
	\caption{The top-1 accuracy of non-structured pruning. We pruned every layer except the first conv., and the sparsity is 0.9. \textbf{Bold} denotes the best results.}
	\begin{tabular}{ccccc|cccc}
	\toprule														
	&	Methods	&	50	&	100	&	500	&	1000/1	&	1000/2	&	1000/3	\\
	\midrule														
	\multirow{2}{*}{0.9}														
	&	CD	&	$48.7_{\pm 0.48}$	&	$53.9_{\pm 0.07}$	&	$59.1_{\pm 0.19}$	&	$60.3_{\pm 0.11}$	&	$61.3_{\pm 0.03}$	&	$61.7_{\pm 0.08}$	\\
	&	$\text{MiR}_{before}$	&	\textbf{52.7}$_{\pm0.53}$	&	\textbf{55.5}$_{\pm0.56}$	&	\textbf{60.9}$_{\pm0.29}$	&	\textbf{63.2}$_{\pm0.11}$	&	\textbf{64.3}$_{\pm0.13}$	&	\textbf{64.7}$_{\pm0.02}$	\\
	\bottomrule	
																				
	\end{tabular}
	\label{app:non-structured}
\end{table*}

\begin{table*}
	\centering
	\caption{Mean and standard deviation of top-1 accuracy (\%) on CIFAR-10. We pruned ResNet-56 with `Res-50\%' used in CD~\cite{Bai_Wu_King_Lyu_2020} and used 1/2/3/5/10/50 samples per class for tuning. Results were reported with five trials. Methods marked with * were copied from CD. \textbf{Bold} denotes the best results.}
	\begin{tabular}{ccccccc}
		\toprule													
		Methods	&	1	&	2	&	3	&	5	&	10	&	50	\\
		\midrule													
		FSKD*	&	$84.26_{\pm 1.42}$	&	$85.79_{\pm 1.31}$	&	$85.99_{\pm 1.29}$	&	$87.53_{\pm 1.06}$	&	$88.15_{\pm 0.71}$	&	$88.70_{\pm 0.55}$	\\
		FitNet*	&	$86.85_{\pm 1.91}$	&	$87.95_{\pm 2.13}$	&	$88.94_{\pm 1.85}$	&	$89.43_{\pm 1.60}$	&	$91.03_{\pm 1.14}$	&	$91.89_{\pm 0.87}$	\\
		ThiNet*	&	$88.40_{\pm 1.26}$	&	$88.76_{\pm 1.18}$	&	$88.95_{\pm 1.19}$	&	$89.54_{\pm 0.84}$	&	$90.36_{\pm 0.76}$	&	$90.89_{\pm 0.49}$	\\
		CP*	&	$88.53_{\pm 1.37}$	&	$88.69_{\pm 1.09}$	&	$88.79_{\pm 0.94}$	&	$89.39_{\pm 0.80}$	&	$89.91_{\pm 0.69}$	&	$90.45_{\pm 0.43}$	\\
		CD*	&	$89.00_{\pm 1.59}$	&	$89.45_{\pm 1.43}$	&	$89.56_{\pm 1.32}$	&	$90.14_{\pm 1.19}$	&	$90.82_{\pm 0.79}$	&	$91.24_{\pm 0.33}$	\\
		\midrule													
		$\text{MiR}_{before}$	& \textbf{89.27}$_{\pm 0.21}$	&	\textbf{90.43}$_{\pm0.12}$	&	\textbf{90.70}$_{\pm0.15}$	&	\textbf{91.14}$_{\pm0.23}$	&	\textbf{91.57}$_{\pm0.14}$	&	\textbf{92.16}$_{\pm0.11}$	\\
		\bottomrule														
	\end{tabular}
	\label{app:cifar10}
\end{table*}

\section{Implementation Details}

In this section, we describe the implementation details of the compared methods.

As mentioned in Sec.~\ref{experiments}, we implemented FSKD for the pruned ResNet models according to its official codes\footnote{https://github.com/LTH14/FSKD}, and we re-ran the official CD codes\footnote{https://github.com/haolibai/Cross-Distillation} to get its results. We only replaced the pruned models while keeping the hyperparameters unchanged. As for BP, we also used SGD as the optimizer, and the initial learning rate, weight decay, and momentum were 1e-3, 1e-4, and $0.9$, respectively. For KD, we also used SGD and used the same learning rate, weight decay, and momentum as those in BP, in which we set the temperature $\tau = 2.0$ and the loss balancing factor was $\alpha = 0.7$.

\section{Extra Results}

In this part, we show results with fewer FLOPs pruned, results with connections pruned, and results on the CIFAR-10 dataset.

\noindent \textbf{Pruning less FLOPs.} As we reported in the main paper, we used `Prune-C Normal' and `Prune-C Residual' to prune the ResNet-34 model (\cf~Tables~\ref{tab:normal_pruning} and~\ref{tab:residual_pruning}, respectively). Here, we illustrated the ResNet-34 results of `Prune-B Normal' (\cf Table~\ref{app:normal_b}) and `Prune-B Residual' (\cf Table~\ref{app:residual_b}) on ILSVRC-2012. As shown in Table~\ref{app:normal_b} and Table~\ref{app:residual_b}, MiR can outperform other methods regardless of pruning a large amount or a small amount of FLOPs, and the gap between $\text{MiR}_{after}$ and $\text{MiR}_{before}$ decreased when pruning less FLOPs.

\noindent \textbf{Unstructured pruning results.} We implemented our MiR to fine-tune the models which were pruned by an unstructured manner (\ie, connection pruning). We used the $\ell_1$-norm weight pruning method to prune the less important weights and compared our MiR with the CD method under $90\%$ weights pruned. Following the same settings in CD, we only pruned the weights in conv. layers, which means the weights in batch normalization layers and fully-connected layers were kept (but we can still update the weights in these layers). And $90\%$ weights here means pruning $90\%$ weights every conv. layer. 

The representation ability was damaged because so much information was lost when pruning $90\%$ weights with smaller $\ell_1$-norm. So we tried a progressive way to prune and fine-tune weights. We pruned $20\%$ weights and then fine-tuned with 400 iterations until we got a $90\%$ pruned sparse model. At last, we fine-tuned the $90\%$ pruned model with 4000 iterations. As for the results of the CD method, we re-ran the official codes and reported the mean and std. of top-1 accuracy.

\noindent \textbf{CIFAR-10 results.} We also conducted ResNet-56 on the CIFAR-10 dataset following the same pruning setting in CD~\cite{Bai_Wu_King_Lyu_2020}. ResNet-56 is a customized model for small datasets like CIFAR-10, and our pre-trained ResNet-56 model has 93.39\%/99.87\% of top-1/top-5 accuracy. We directly compared our $\text{MiR}_{before}$ with FSKD~\cite{Li_2020_CVPR}, FitNet~\cite{Romero_fitnets_2015}, ThiNet~\cite{luo_thinet_2017}, CP~\cite{he_channel_2017} and CD~\cite{Bai_Wu_King_Lyu_2020}. As for our MiR, no extra augmentation was used except for random horizontal flip, and training settings were the same as mentioned in Sec.~\ref{experiments}.

In Table~\ref{app:cifar10}, we used 1/2/3/5/10/50 samples per class to fine-tune the pruned models and reported the mean and std. of top-1 accuracy under five independent trials. Results marked with * were copied from CD. As shown in Table~\ref{app:cifar10}, our MiR worked well in all cases.

\end{appendix}

\end{document}